%% file: main.tex
\documentclass[runningheads]{llncs}
\usepackage[T1]{fontenc}
\usepackage{graphicx}
\usepackage{amssymb}
\usepackage{bm}
\usepackage{mathtools}
\usepackage{booktabs}
\usepackage{enumitem}
\usepackage{multirow}
\usepackage{capt-of}
\usepackage{hyperref}
\hypersetup{colorlinks=true,citecolor=blue}
\usepackage{color}

\newcommand{\Norm}[1]{\left\lVert #1 \right\rVert}
\begin{document}
\title{Shape-guided Gaussian Splatting for Sparse-View X-ray 3D Reconstruction}

\author{Pranav Poudel \inst{1, 3}$^{*}$ \and
Florence Dell'Aniello Picard\inst{1,3} \and
Nairouz Shehata \inst{1, 3} \and
Frédéric Lavoie \inst{2} \and
 Herve Lombaert \inst{1, 3}
}

\authorrunning{P. Poudel et al.}

\institute{Polytechnique Montréal, Canada\\
\and
CHUM - University of Montreal Hospital, Canada\\
\and
Mila - Quebec AI Institute, Canada
}

\renewcommand{\thefootnote}{*}
\footnotetext[1]{Correspondence to: Pranav Poudel <pranav.poudel@etud.polymtl.ca>}
\renewcommand{\thefootnote}{\arabic{footnote}} 

\maketitle              
\begin{abstract}
Sparse-view X-ray 3D reconstruction is essential for reducing radiation exposure, but recovering a density field from a handful of X-ray projections is severely ill-posed. Recently, 3D Gaussian Splatting has achieved state-of-the-art performance in sparse-view reconstruction by representing the volume using explicit, optimized primitives, but it requires dozens of projected views. With fewer views, reconstruction quality degrades severely since the explicit primitives are optimized freely without any anatomical information. Anatomical structures, in contrast, share similar geometry and density across a population. Their variations are bounded within a limited range that statistical shape models can capture. This paper proposes a shape-guided Gaussian splatting framework for sparse-view X-ray 3D reconstructions. Our contribution lies in driving Gaussian positions toward anatomically valid configurations, alongside atlas-based density regularization. Our method ensures anatomically consistent reconstruction and improves PSNR by 2.83 dB over a state-of-the-art Gaussian splatting baseline with as few as 5 views.  Code Available: \url{https://github.com/polyshape-lab/ShapeGuidedGaussian}
\end{abstract}
\input{introduction}

\input{method}
\input{experiments}
\input{results}
\input{conclusion}

\begin{credits}
\subsubsection{\ackname}
This project is supported by the NSERC Alliance Advantage grant in partnership with Eiffel Medtech Inc. We acknowledge the Digital Research Alliance of Canada for providing computational resources, and the New Mexico Decedent Image Database (NMDID) for the CT imaging data.
 
\subsubsection{\discintname}
This work has received research funding from Eiffel Medtech Inc. Dr. Lavoie is the founder of Eiffel Medtech Inc., and the other authors have no competing interests to declare.
\end{credits}
\bibliographystyle{splncs04}
\bibliography{ref}
\end{document}

%% file: introduction.tex
\section{Introduction}
3D reconstructions of human anatomy are relevant to clinical applications, particularly in orthopedics, such as preoperative planning, surgical navigation, and patient-specific instrumentation~\cite{d2022use,RAMBANI201450}. X-ray imaging is widely used in routine clinical practice. However, recovering  3D geometry from a few 2D X-ray projections is an ill-posed problem. CT imaging can reconstruct a high-resolution 3D volume by acquiring hundreds of projections and inverting the Radon transform~\cite{kak2001principles}, but at a much higher radiation dose than standard X-ray imaging~\cite{cao2022ct}. Reducing the number of projections lowers this dose and eases acquisition in clinical settings such as the operating room. Prior work on recovering 3D anatomy from X-ray projections falls into two families: surface reconstruction~\cite{baka20112d,8705271,hosseinian20153d,karade20153d} and volumetric reconstruction~\cite{r2_gaussian,zha2022naf,cai2024structure,Feldkamp:84,andersen1984simultaneous}. Surface reconstruction methods recover the 3D boundary using a statistical shape model, but rely on features such as accurate contours and landmarks of its boundary. Volumetric reconstructions recover a density field, i.e. the X-ray attenuation at each point in the object,  directly from X-ray projections, but require dozens of projections. We focus on volumetric reconstructions, often referred to as tomographic reconstructions.

Traditional tomographic reconstruction methods~\cite{Feldkamp:84,andersen1984simultaneous} produce high-quality reconstructions with hundreds of projections, but quality degrades severely under sparse views, leading to streak artifacts and loss of structural detail. More recently, NeRF-based methods~\cite{zha2022naf,cai2024structure}, and 3D Gaussian splatting methods~\cite{cai2024radiative,r2_gaussian} have drawn interest for volumetric reconstruction. NeRF-based methods model the density field with multilayer perceptrons, whereas Gaussian splatting uses explicit Gaussian primitives; both are trained with photometric loss. Among these, $R^{2}$-Gaussian~\cite{r2_gaussian} achieves state-of-the-art performance in sparse-view settings ($25\text{--}75$ views) by addressing the integration bias of 3DGS~\cite{kerbl20233d}. However, both families predict the density field solely from projections, with no anatomical prior during optimization. Under extremely sparse views, reconstructions become severely ill-posed, with many density fields that can project to the same view. These methods then overfit the available views, producing ill-formed boundaries.

Human anatomy shares similar geometry across individuals, with variation bounded within a limited range~\cite{cootes1995active}. Furthermore, organs have extensive homogeneous regions within but change sharply at their boundaries. Statistical shape models capture such regularities and are widely used in a variety of medical imaging applications~\cite{heimann2009statistical}. Shape models based on Gaussian Splatting have been used in computer vision, notably for faces and human avatars~\cite{zhao2024psavatar,qian2024gaussianavatars}. However, such methods require large numbers of images from video clips, which is often infeasible in a medical setting. The bounded variation captured by a shape model can serve as a strong prior that can meet the requirements of many 2D projections in medical settings. Integrating shape models into Gaussian splatting remains underexplored for medical imaging. Shape-based priors have been used to constrain 3D reconstruction from a few projections, through learned structure priors~\cite{10.1007/978-3-031-43999-5_66} and atlas-based 2D/3D registration~\cite{van2022deep}. To our knowledge, no existing work binds Gaussian primitives to a shape model for X-ray 3D reconstruction.

To overcome the lack of anatomical constraints in existing sparse-view 3D reconstruction methods, a shape-guided Gaussian splatting framework is proposed that uses population shape and density information to guide reconstruction. Our guidance is applied at two stages. \textit{During initialization}, Gaussian positions and orientations are seeded from the shape template, and Gaussian densities are seeded from the population-average density from the intensity image. \textit{During optimization}, Gaussian positions are collectively deformed within an anatomically plausible region via a shape prior, while a density prior constrains the reconstructed volume to statistical variations in density. Our method substantially outperforms the state-of-the-art baseline when reconstructing femoral shapes under extremely sparse views. Our contributions are as follows:
\begin{itemize}
    \item We propose a shape-guided Gaussian splatting framework that binds the positions of Gaussian primitives to a shape model, constraining the geometry to anatomically valid configurations.
    \item We incorporate guidance in two stages: Initialization and Optimization. We seed orientation and appearance of primitives from the population mean and regularize the reconstruction towards population statistics.
    \item We conduct experiments on femoral data from NMDID~\cite{Edgar2020NMDID} and demonstrate consistent improvements over the state-of-the-art baseline under extreme sparsity as low as 5-view reconstruction.  
\end{itemize}

%% file: method.tex
\begin{figure}[t]
    \centering
    \includegraphics[width=\linewidth]{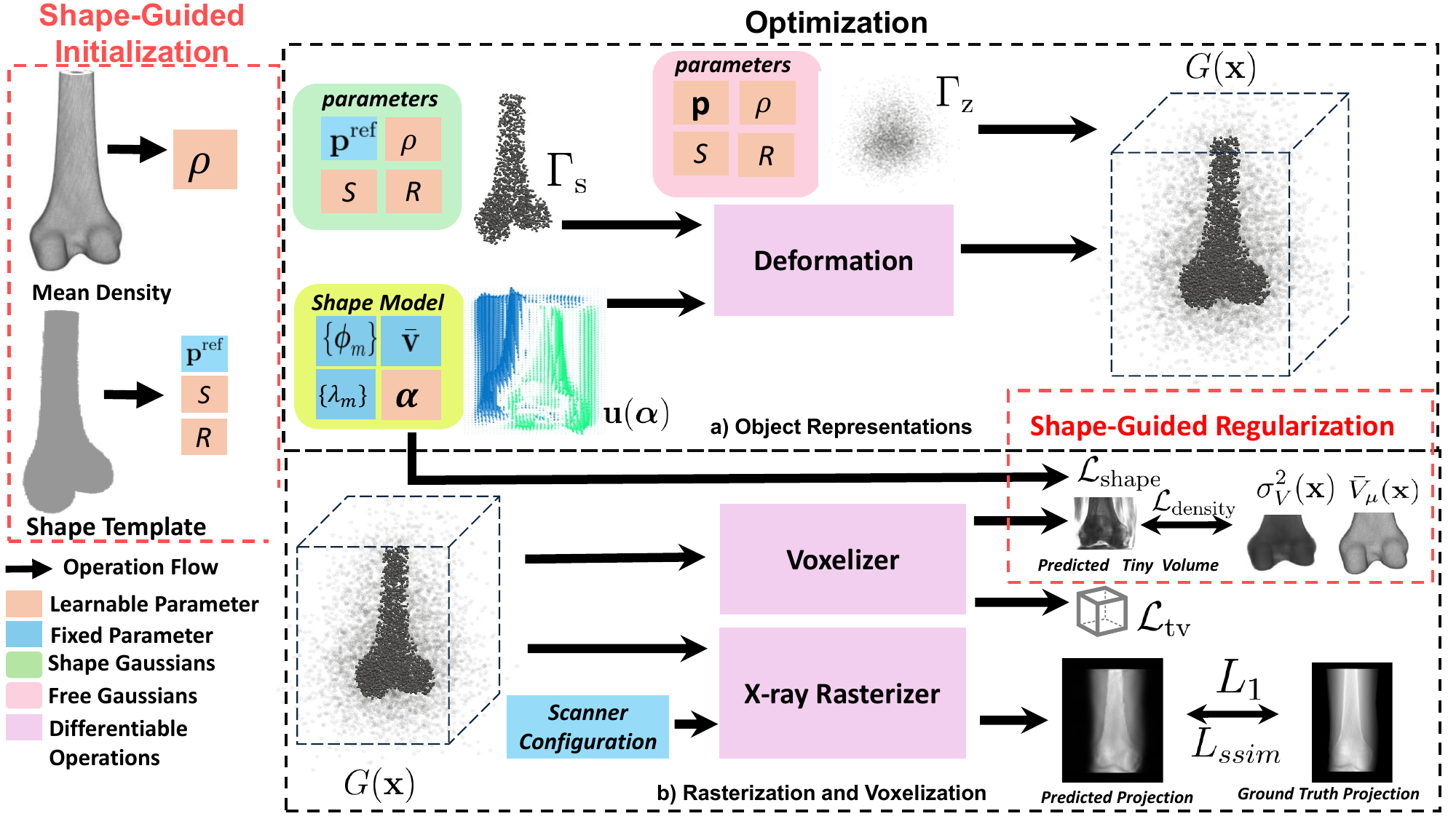}
    \caption{\textbf{Overview of our framework.} Shape Gaussians ($\Gamma_{\text{s}}$) are initialized from a template and population mean density, seeding reference position $\boldsymbol{\mathbf{p}}^{\text{ref}}$, scaling $S$, rotation $R$ and density $\rho$. Positions deform collectively via a displacement field $\mathbf{u}(\boldsymbol{\alpha})$ generated by the shape coefficients $\boldsymbol{\alpha}$, while scaling, rotation, and density are updated independently. The full Gaussian primitives set ($G =\Gamma_{\text{s}} \cup \Gamma_{\text{z}}$ ) are projected by the X-ray rasterizer, and all parameters are optimized with photometric ($\mathcal{L}_1$, $\mathcal{L}_\text{ssim}$) and regularization  ($\mathcal{L}_\text{tv}$, $\mathcal{L}_\text{density}$, $\mathcal{L}_\text{shape}$) losses.}
    \label{fig:pipeline}
\end{figure}
\section{Method}
The proposed method guides Gaussian splatting using population priors to ensure that reconstruction remains anatomically plausible. As shown in Fig.~\ref{fig:pipeline}, the guidance occurs in two stages: Initialization and Optimization. In this section, the object representation is introduced first (Sec.~\ref{subsec:object_representation}), followed by the shape model that guides the Gaussian primitives (Sec.~\ref{subsec:shape_model}). The Parameterization and Initialization of Gaussian primitives using the shape model are elaborated (Sec.~\ref{subsec:param_init}), followed by their optimization (Sec.~\ref{subsec:optim}).
\subsection{Object Representations}
\label{subsec:object_representation}
The 3D volume can be represented as a collection of $N$ radiative Gaussian primitives~\cite{r2_gaussian} $\{G_i\}_{i=1}^{N}$ each defined as, 
\begin{equation}
    G_i(\mathbf{x}) = \rho_i \cdot \exp\!\left( -\tfrac{1}{2} (\mathbf{x} - \boldsymbol{\mathbf{p}}_i)^{\top} \Sigma_i^{-1} (\mathbf{x} - \boldsymbol{\mathbf{p}}_i) \right),
    \label{eq:formulation}
\end{equation}
with position $\boldsymbol{\mathbf{p}}_i \in \mathbb{R}^3$, covariance $\Sigma_i$
, and central density $\rho_i \in \mathbb{R}_{\geq 0}$. For optimization purposes~\cite{kerbl20233d}, the covariance matrix is decomposed as $\Sigma_i = R_i \, S_i \, S_i^{\top} R_i^{\top}$, where $R_i$ is the rotation and $S_i$ is the scaling matrix. This collection of primitives is partitioned into two sets: a shape-guided set $\Gamma_{\text{s}}$, representing the anatomy of interest that will be guided by the shape model, and a free set $\Gamma_{\text{z}}$ representing the rest of the volume. Thus, primitives in $\Gamma_{\text{z}}$ have position, covariance, and density all as free learnable parameters, whereas Gaussians in $\Gamma_{\text{s}}$ have their position driven by a shape model. Since Gaussian primitives combine additively, the density field at ray position $\mathbf{x}$ is defined as $\mu(\mathbf{x}) = \sum_{i=1}^{N} G_i(\mathbf{x}).$ A differentiable rasterizer renders a projection $I_r$ as an analytic line integral of the attenuation field along each ray, and the voxelizer queries the field on a grid to produce a density volume for 3D reconstruction~\cite{r2_gaussian}.
\subsection{Shape Model}
\label{subsec:shape_model}
The 3D geometry of an object is represented in a low-dimensional shape space learned from a population of $P$ segmentations. Each subject in the population is registered to a common template using diffeomorphic registration~\cite{arsigny2006log}, yielding a per-subject velocity field $\mathbf{v}$. A PCA over these velocity fields gives a mean velocity field $\bar{\mathbf{v}}$ and orthonormal modes $\{\boldsymbol{\phi}_m\}$ with variance $\{\lambda_m \}$, so that a velocity field is generated from $M$ shape coefficients, $\boldsymbol{\alpha} \in \mathbb{R}^{M}$,

\begin{equation}
    \mathbf{v}(\boldsymbol{\alpha}) = \bar{\mathbf{v}} + \sum_{m=1}^{M} \alpha_m\, \sigma_m\, \boldsymbol{\phi}_m, \qquad \sigma_m = \sqrt{\lambda_m}
    \label{eq:shape_space}
\end{equation}

\noindent where standardizing by $\sigma_m$ induces the prior $\boldsymbol{\alpha} \sim \mathcal{N}(\mathbf{0}, I)$. The velocity field $\mathbf{v}(\boldsymbol{\alpha})$ is integrated to obtain a displacement field $\mathbf{u}(\boldsymbol{\alpha})$ which is used to deform shape Gaussian primitives~(Eq.\ref {eq:displacement}). From the same registered population, per-voxel mean $\bar{V}_\mu(\mathbf{x})$ and variance $\sigma_V^2(\mathbf{x})$ of the density are computed to form a density atlas. This atlas is used for density initialization and confidence weight in the density prior loss.

\subsection{Parameterization and Initialization}
\label{subsec:param_init}
The core of the proposed method is to bind the Gaussian primitives to the shape model. Each $G_j(\mathbf{x})\in\Gamma_{\text{s}}$ sits at a fixed reference center $\boldsymbol{\mathbf{p}}_j^{\text{ref}}$ on the template. Instead of moving freely, it is carried by a deformation induced by the shape coefficients $\boldsymbol{\alpha}$, where the resulting displacement field $\mathbf{u}(\boldsymbol{\alpha})$ is sampled at the reference position. So, the position of each primitive in $\Gamma_{\text{s}}$ is given by,

\begin{equation}
    \boldsymbol{\mathbf{p}}_j = \boldsymbol{\mathbf{p}}_j^{\text{ref}} + \mathbf{u}_j(\boldsymbol{\alpha}).
    \label{eq:displacement}
\end{equation}
In contrast, the scaling $S_j$, rotation $R_j$ and density $\rho_j$ that encode orientation and appearance remain independent parameters optimized directly. This will result in Gaussian primitives being in anatomically valid positions, while their appearance and orientation remain adaptive to the available projections.  Thus, the learnable set of shape Gaussian primitives is: $
    \Theta = \big\{\, \boldsymbol{\alpha} \,;\, \{S_j\}, \{R_j\}, \{\rho_j\} \,\big\}$.

Prior work initializes Gaussian primitives from a low-quality volume reconstructed using FDK~\cite{Feldkamp:84}. Under extreme sparsity, this reconstruction becomes severely degraded, giving an unreliable starting point for optimization. Instead, the orientation of the shape Gaussian primitives is initialized from the template, and the density is initialized from the population mean density. Reference centers ${\{\mathbf{p}_j^{\text{ref}}\}_{j=1}^{N_s}}$, where $N_s = |\Gamma_{\text{s}}|$, are obtained by boundary-weighted sampling of the template, placing denser primitives near the boundary where sharp transitions require more primitives to represent. Scaling parameters are seeded from the nearest-neighbour spacing of reference positions~\cite{r2_gaussian,kerbl20233d}. However, unlike their isotropic identity-rotation initialization, the surface primitives are made anisotropic by thinning primitives along the normal by a factor $\tau \in (0, 1)$, and orienting them in order to align their thinned axis with the normal of the template surface. This reorientation causes the Gaussian primitives to be flat across thin surfaces~\cite{guedon2023sugar}. Densities are initialized so that the accumulated field matches the per-voxel population mean density. Since primitives overlap and combine additively, a single primitive's density cannot be read directly from the target. To account for this, all densities are set to one and voxelized to measure how much overlap occurs at each location using Eq.~\ref{eq:overlap}
\begin{equation}
    n(\mathbf{x}) = \sum_{i \in \Gamma_{\text{s}}} \exp\!\left( -\tfrac{1}{2}(\mathbf{x}-\mathbf{p}_i)^{\top}\Sigma_i^{-1}(\mathbf{x}-\mathbf{p}_i) \right).
    \label{eq:overlap}
\end{equation}
Each density is then set to atlas mean density $\bar{V}_\mu$ at the Gaussian primitives reference location $\boldsymbol{\mathbf{p}}_j^{\text{ref}}$, divided by the local overlap defined in Eq.~(\ref{eq:overlap}) as:
\begin{equation}
\rho_j^{\text{init}} = \frac{\bar{V}_\mu(\mathbf{p}_j^{\text{ref}})}{\max\!\big(n(\mathbf{p}_j^{\text{ref}}),\, 1\big)}.
\end{equation}
Hence, the overlapping Gaussians sum to the correct mean density.
\subsection{Optimization Strategy}
\label{subsec:optim}
We optimize Gaussian primitives using gradient descent by minimizing the total objective $L$ (Eq.~\ref{eq:objective}). Besides combining photometric losses ($\mathcal{L}_1$ and D-SSIM $\mathcal{L}_\text{ssim}$~\cite{wang2004image}) and a 3D total-variation prior ($\mathcal{L}_\text{tv}$~\cite{RUDIN1992259}), two regularization terms are introduced specific to the shape-guided setting: a shape prior on the shape coefficients and a density prior on the reconstructed volume. Shape modes are standardized so $\boldsymbol{\alpha} \sim \mathcal{N}(\mathbf{0}, I)$, with plausible shapes lying near the origin of shape space. Hence, deviation is penalized as,
\begin{equation}
    \mathcal{L}_{\text{shape}} = \Norm{\boldsymbol{\alpha}}_2^2,
    \label{eq:mahal}
\end{equation}
which is the squared Mahalanobis distance of $\boldsymbol{\alpha}$ under shape prior. This keeps the deformed primitives within anatomically plausible shapes. A density prior pulls the reconstructed volume towards the population mean. But the population mean is only reliable where its value is consistent across subjects. On a randomly sampled sub-volume grid $\Omega$, only shape Gaussian primitives are queried to obtain volume $\hat{V}$ and compute the density prior as:
\begin{equation}
    w(\mathbf{x}) = \exp\!\left( -\frac{\sigma_V^2(\mathbf{x})}{s^2} \right),
\qquad
\mathcal{L}_{\text{density}} = \frac{1}{|\Omega|} \sum_{\mathbf{x} \in \Omega} w(\mathbf{x}) \,\big( \hat{V}(\mathbf{x}) - \bar{V}_\mu(\mathbf{x}) \big)^2,
\label{eq:density_reg}
\end{equation}
where $s$  controls the fall-off of the confidence weight $w$. Confidence is high where the population is consistent and low where it is highly variable. Hence, the final objective function becomes,
\begin{equation}
\mathcal{L} = \mathcal{L}_1 + \lambda_{\text{ssim}}\mathcal{L}_{\text{ssim}} + \lambda_{\text{tv}}\mathcal{L}_{\text{tv}} + \lambda_{\text{shape}}\mathcal{L}_{\text{shape}} + \lambda_{\text{density}}\mathcal{L}_{\text{density}}.
\label{eq:objective}
\end{equation}

%% file: experiments.tex
\section{Experimental Setup}
\textbf{Datasets and Shape Model:} We conduct experiments on synthetic datasets by leveraging 758 CT scans from NMDID~\cite{Edgar2020NMDID}, resampled into $(1,1,1)$ mm spacing. TotalSegmentator~\cite{wasserthalTotalSegmentatorRobustSegmentation2023} was used to generate femoral segmentations, and ANTs~\cite{tustisonANTsXEcosystemQuantitative2021} was used for rigid alignment of all subjects to a common reference. Of the 758 subjects, 106 were held out: 100 to build a shape model and 6 for reconstruction. Since reconstruction is per-subject optimization, 6 subjects constitute 6 independent evaluations. The remainder was used to train VoxelMorph~\cite{balakrishnan2019tmi} for diffeomorphic registration. The trained model then co-registers the 100 subjects onto a common template, yielding a stationary velocity field for each subject. PCA over these fields was applied~\cite{turk1991eigenfaces}, retaining $M$ = 35 components that capture 95\% of variance. The voxelwise mean and variance over the same population were calculated for density initialization and the confidence weight in the density prior loss.  We then use TIGRE~\cite{biguri2016tigre} to synthesize X-ray projections over the $0\text{--}180^{\circ}$ range, incorporating Compton scattering and electronic noise. To evaluate whether shape prior improves reconstruction quality under extreme sparsity, we set the number of views to 5 and 10, far sparser than the $25\text{--}75$ views settings of prior work~\cite{r2_gaussian}.

\noindent \textbf{Baseline and metrics}
We compare our method against $R^2$-Gaussian~\cite{r2_gaussian}, a state-of-the-art 3D reconstruction method based on Gaussian splatting. We use PSNR and SSIM~\cite{wang2004image} to assess reconstruction quality, with PSNR calculated in 3D and SSIM averaged over 2D slices along 3 axes. 

\noindent \textbf{Implementation Details: } Both our method and baseline are trained for 30K iterations with Adam~\cite{kingma2014adam} on a single A100 GPU. For photometric losses, we use the default hyperparameters of~\cite{r2_gaussian}. Proposed loss weights were selected as $\lambda_{\text{shape}} = 0.0001$, $\lambda_{\text{density}} = 0.5$, confidence fall-off $s=0.05$, thinning factor $\tau = 0.5$ empirically on a single subject. Initial learning rates are 0.1($\boldsymbol{\alpha}$), 0.01($\rho$), 0.005($\mathbf{S}$) and 0.01($\mathbf{R}$), each decaying exponentially to 10\% of its initial value, except $\boldsymbol{\alpha}$, which decays to 1\% by the midpoint of training. The shape coefficients are frozen during a short warm-up of 50 iterations to stabilize the appearance. We initialize $N_s$ = 35K shape Gaussians and $N_z$ = 15K free Gaussians (50K total matching baseline).

%% file: results.tex
\begin{figure}[!t]
    \centering
    \includegraphics[width=\linewidth]{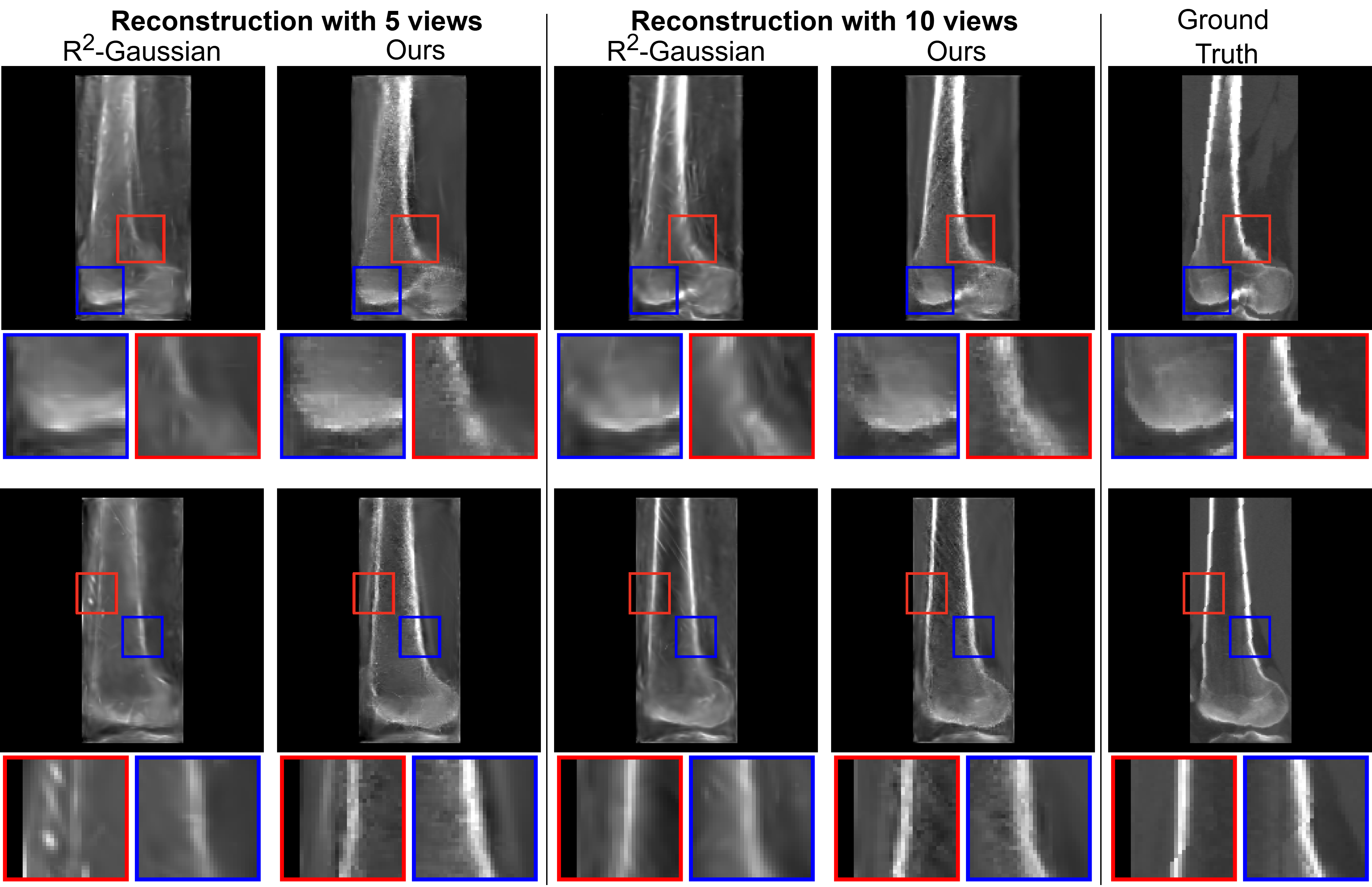}
    \caption{Qualitative comparison slices shown across two axes (top: coronal; bottom: sagittal). Compared to $R^2$-Gaussian, our method recovers sharper cortical boundaries and finer internal structure, closer to the ground truth.}
    \label{fig:qual}
\end{figure}
\section{Results}
\input{tables/3d_results}
Our experiments evaluate whether incorporating anatomical priors into Gaussian splatting improves reconstruction quality in extremely sparse-view settings. Table~\ref{tab:3d} shows the PSNR and SSIM for the baseline and our method across six subjects (\textbf{S1}--\textbf{S6}) at 5 and 10 views. Our results show that our method significantly outperforms the baseline, and the improvement is consistent across all subjects and both views. For example, our guidance improves the average 3D PSNR by +2.83 dB (34.55 $\rightarrow$ 37.38) and 3D SSIM by +0.022 (0.937 $\rightarrow$ 0.959) at 5 views, while at 10 views the gains are smaller: +1.46 dB (38.89 $\rightarrow$ 40.35) and +0.012 (0.959 $\rightarrow$ 0.971). The gap narrows as the number of views increases because the projections themselves provide more constraints. 
We further examine whether incorporating the shape prior yields sharper, anatomically plausible boundaries than the baseline. Fig.~\ref{fig:qual} shows reconstructions of \textbf{S3} at 5 and 10 views along two different axes. At 5 views, we observe that the $R^2$-Gaussian produces blurred boundaries, whereas our method recovers sharp boundaries and coherent structure. In the highlighted insets, we observe that the cortical boundary of the $R^2$-Gaussian is smeared and spreads into the surrounding areas. In contrast, our method remains sharp and closely follows the ground-truth contour. 
\input{tables/ablation_study}
\noindent \textbf{Ablation Study:} We perform an ablation study to show the contribution of each proposed component. Table~\ref{tab:ablation} and Fig.~\ref{fig:ablation} show the contribution of each component on 5-view settings. Starting from density initialization alone, where all primitives are independent, adding the shape prior by binding primitives to the shape model yields a large improvement of +4.33 dB. Fig.~\ref{fig:ablation} illustrates that with independent primitives, the rendered training-view angle projection closely matches the ground-truth projection, but the reconstructed volume does not. This is because only primitives along the observed rays are effectively optimized. This results in a mottled, inconsistent interior despite having an initial boost. Adding the shape prior resolves this because all the primitives are deformed collectively. The density prior adds a further, smaller gain, refining the density.

%% file: tables/3d_results.tex
\begin{table}[!htbp]
\centering
\caption{Quantitative 3D results across six test subjects (\textbf{S1}--\textbf{S6}). Best in bold}
\label{tab:3d}
\footnotesize
\setlength{\tabcolsep}{2pt}
\renewcommand{\arraystretch}{0.9}
\begin{tabular}{cccccccccc}
\toprule
\textbf{$\#$ of Views} & \textbf{Metric} & \textbf{Method} & \textbf{S1} & \textbf{S2} & \textbf{S3} & \textbf{S4} & \textbf{S5} & \textbf{S6} & \textbf{Avg.} \\
\midrule
\multirow{4}{*}{5}
  & \multirow{2}{*}{PSNR$\uparrow$}
    & R2GS  & 34.44 & 33.03 & 35.01 & 35.43 & 34.84 & 34.56 & 34.55 \\
  & & Ours  & \textbf{37.71} & \textbf{35.69} & \textbf{37.93} & \textbf{37.48} & \textbf{37.41} & \textbf{38.06} & \textbf{37.38} \\
\cmidrule{2-10}
  & \multirow{2}{*}{SSIM$\uparrow$}
    & R2GS  & 0.937 & 0.925 & 0.943 & 0.943 & 0.937 & 0.936 & 0.937 \\
  & & Ours  & \textbf{0.961} & \textbf{0.948} & \textbf{0.965} & \textbf{0.960} & \textbf{0.957} & \textbf{0.961} & \textbf{0.959} \\
\midrule
\multirow{4}{*}{10}
  & \multirow{2}{*}{PSNR$\uparrow$}
    & R2GS  & 39.08 & 38.05 & 39.51 & 39.58 & 38.68 & 38.44 & 38.89 \\
  & & Ours  & \textbf{40.48} & \textbf{39.77} & \textbf{40.77} & \textbf{40.97} & \textbf{40.14} & \textbf{39.98} & \textbf{40.35} \\
\cmidrule{2-10}
  & \multirow{2}{*}{SSIM$\uparrow$}
    & R2GS  & 0.961 & 0.955 & 0.965 & 0.964 & 0.956 & 0.954 & 0.959 \\
  & & Ours  & \textbf{0.972} & \textbf{0.968} & \textbf{0.974} & \textbf{0.974} & \textbf{0.968} & \textbf{0.967} & \textbf{0.971} \\
\bottomrule
\end{tabular}
\end{table}

%% file: tables/ablation_study.tex
\begin{figure}[!t]
\centering
\begin{minipage}[c]{0.62\linewidth}
\centering
\includegraphics[width=0.8\linewidth]{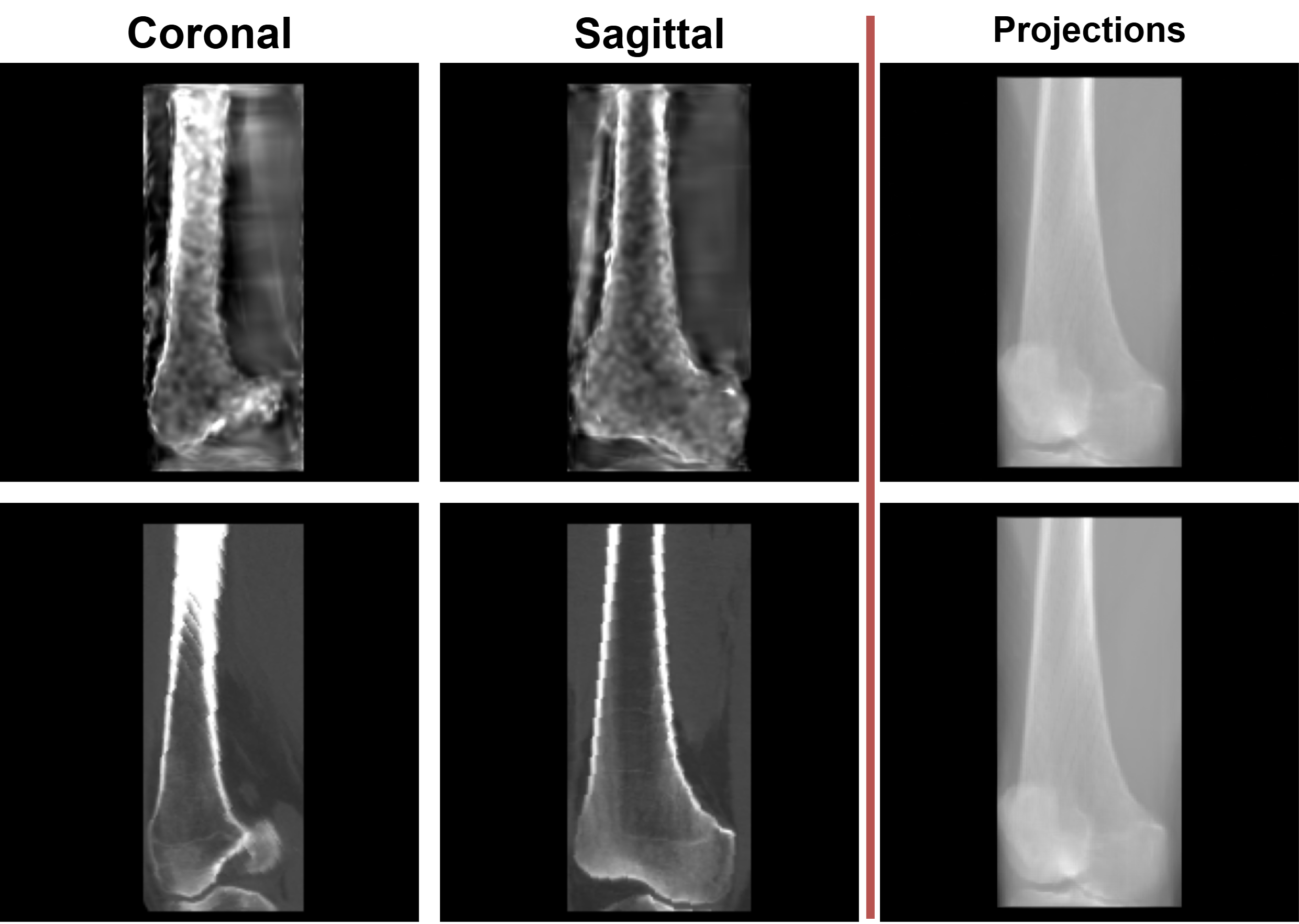}
\caption{\textbf{Ablation Study on Initialization}: Free Gaussian (top) at 5 views vs. ground truth (bottom). Matched Projection at training view, yet the reconstructed volume does not match ground truth.}
\label{fig:ablation}
\end{minipage}
\hfill
\begin{minipage}[c]{0.36\linewidth}
\centering
\captionof{table}{Ablation on 5-view reconstruction, showing the contribution of each proposed component.}
\label{tab:ablation}
\setlength{\tabcolsep}{0.5pt}
\begin{tabular}{lcc}
\toprule
\textbf{Method} & \textbf{PSNR}$\uparrow$ & \textbf{SSIM}$\uparrow$ \\
\midrule
Density Init.                  & 32.93 & 0.925 \\
+ $\lambda_\text{shape}$       & 37.26 & 0.958 \\
+ $\lambda_\text{density}$     & \textbf{37.38} & \textbf{0.959} \\
\bottomrule
\end{tabular}
\end{minipage}%
\end{figure}



%% file: conclusion.tex
\section{Conclusion}
We presented a shape-guided Gaussian splatting framework for sparse-view X-ray 3D reconstruction. This framework constrains the locations of Gaussian primitives to anatomically valid configurations using a shape model and leverages population statistics for regularization. Furthermore, our framework initializes the Gaussian parameters from population statistics, giving an anatomically valid starting point for optimization. Our method improves the PSNR by 2.83 dB and SSIM by 0.022 at only 5 views with respect to competitive state-of-the-art methods under extreme sparsity. Our method is, in theory, not limited to our tested femur, suggesting potential extensions to other anatomical structures.

%% file: ref.bib
@inproceedings{r2_gaussian,
  title={R$^2$-Gaussian: Rectifying Radiative Gaussian Splatting for Tomographic Reconstruction},
  author={Ruyi Zha and Tao Jun Lin and Yuanhao Cai and Jiwen Cao and Yanhao Zhang and Hongdong Li},
  booktitle = {Advances in Neural Information Processing Systems (NeurIPS)},
  year={2024}
}

@article{RUDIN1992259,
title = {Nonlinear total variation based noise removal algorithms},
journal = {Physica D: Nonlinear Phenomena},
volume = {60},
year = {1992},
author = {Leonid I. Rudin and Stanley Osher and Emad Fatemi},
}

@article{balakrishnan2019tmi,
    title={VoxelMorph: A Learning Framework for Deformable Medical Image Registration},
    author={Balakrishnan, Guha and Zhao, Amy and Sabuncu, Mert and Guttag, John and Adrian V. Dalca},
    journal={IEEE Transactions on Medical Imaging},
    volume={38},
    year={2019}
}

@article{biguri2016tigre,
  title={TIGRE: a MATLAB-GPU toolbox for CBCT image reconstruction},
  author={Biguri, Ander and Dosanjh, Manjit and Hancock, Steven and Soleimani, Manuchehr},
  journal={Biomedical Physics \& Engineering Express},
  volume={2},
  year={2016},
  publisher={IOP Publishing}
}

@article{turk1991eigenfaces,
  title={Eigenfaces for recognition},
  author={Turk, Matthew and Pentland, Alex},
  journal={Journal of Cognitive Neuroscience},
  volume={3},
  year={1991},
  publisher={MIT Press One Rogers Street, Cambridge, MA 02142-1209, USA journals-info~…}
}

@inproceedings{kingma2014adam,
  title={Adam: A method for stochastic optimization},
  author={Kingma, Diederik P and Ba, Jimmy},
  booktitle={International Conference on Learning Representations (ICLR)},
  year={2015}
}

@article{Feldkamp:84,
author = {L. A. Feldkamp and L. C. Davis and J. W. Kress},
journal = {J. Opt. Soc. Am. A},
publisher = {Optica Publishing Group},
title = {Practical cone-beam algorithm},
volume = {1},
year = {1984},
}

@inproceedings{zha2022naf,
  title={NAF: neural attenuation fields for sparse-view CBCT reconstruction},
  author={Zha, Ruyi and Zhang, Yanhao and Li, Hongdong},
  booktitle={International Conference on Medical Image Computing and Computer-Assisted Intervention (MICCAI)},
  year={2022},
}

@inproceedings{cai2024structure,
  title={Structure-aware sparse-view x-ray 3d reconstruction},
  author={Cai, Yuanhao and Wang, Jiahao and Yuille, Alan and Zhou, Zongwei and Wang, Angtian},
  booktitle={IEEE/CVF Conference on Computer Vision and Pattern Recognition (CVPR)},
  year={2024}
}

@inproceedings{cai2024radiative,
  title={Radiative gaussian splatting for efficient x-ray novel view synthesis},
  author={Cai, Yuanhao and Liang, Yixun and Wang, Jiahao and Wang, Angtian and Zhang, Yulun and Yang, Xiaokang and Zhou, Zongwei and Yuille, Alan},
  booktitle={European Conference on Computer Vision (ECCV)},
  year={2024},
}

@article{cootes1995active,
  title={Active shape models-their training and application},
  author={Cootes, Timothy F and Taylor, Christopher J and Cooper, David H and Graham, Jim},
  journal={Computer Vision and Image Understanding},
  volume={61},
  year={1995},
  publisher={Elsevier}
}

@inproceedings{qian2024gaussianavatars,
  title={Gaussianavatars: Photorealistic head avatars with rigged 3d gaussians},
  author={Qian, Shenhan and Kirschstein, Tobias and Schoneveld, Liam and Davoli, Davide and Giebenhain, Simon and Nie{\ss}ner, Matthias},
  booktitle={IEEE/CVF Conference on Computer Vision and Pattern Recognition (CVPR)},
  year={2024}
}

@article{zhao2024psavatar,
  title={Psavatar: A point-based shape model for real-time head avatar animation with 3d gaussian splatting},
  author={Zhao, Zhongyuan and Bao, Zhenyu and Li, Qing and Qiu, Guoping and Liu, Kanglin},
  journal={IEEE Transactions on Visualization and Computer Graphics},
  year={2026},
  volume={32},
  publisher={IEEE}
}

@misc{Edgar2020NMDID,
  author       = {Edgar, H. J. H. and Daneshvari Berry, S. and Moes, E. and Adolphi, N. L. and Bridges, P. and Nolte, K. B.},
  title        = {New {M}exico {D}ecedent {I}mage {D}atabase},
  howpublished = {Office of the Medical Investigator, University of New Mexico},
  year         = {2020}
}

@article{wang2004image,
  title={Image quality assessment: from error visibility to structural similarity},
  author={Wang, Zhou and Bovik, Alan C and Sheikh, Hamid R and Simoncelli, Eero P},
  journal={IEEE Transactions on Image Processing},
  volume={13},
  year={2004},
  publisher={IEEE}
}

@article{d2022use,
  title={The use of computerized tomography scans in elective knee and hip arthroplasty—what do they tell us and at what risk?},
  author={D’Amore, Taylor and Klein, Gregg and Lonner, Jess},
  journal={Arthroplasty Today},
  volume={15},
  year={2022},
  publisher={Elsevier}
}

@book{kak2001principles,
  title={Principles of computerized tomographic imaging},
  author={Kak, Avinash C and Slaney, Malcolm},
  year={2001},
  publisher={SIAM}
}

@article{kerbl20233d,
      author       = {Kerbl, Bernhard and Kopanas, Georgios and Leimk{\"u}hler, Thomas and Drettakis, George},
      title        = {3D Gaussian Splatting for Real-Time Radiance Field Rendering},
      journal      = {ACM Transactions on Graphics},
      volume       = {42},
      year         = {2023},
}

@article{cao2022ct,
  title={CT scans and cancer risks: a systematic review and dose-response meta-analysis},
author={Cao, Chun-Feng and Ma, Kun-Long and Shan, Hua and Liu, Tang-Fen and Zhao, Si-Qiao and Wan, Yi and Zhang, Jun and Wang, Hai-Qiang},  
journal={BMC Cancer},
  volume={22},
  year={2022},
  publisher={Springer}
}

@article{RAMBANI201450,
title = {Computer assisted navigation in orthopaedics and trauma surgery},
journal = {Orthopaedics and Trauma},
volume = {28},
year = {2014},
author = {Rohit Rambani and Mathew Varghese},

}

@article{heimann2009statistical,
  title={Statistical shape models for 3D medical image segmentation: a review},
  author={Heimann, Tobias and Meinzer, Hans-Peter},
  journal={Medical Image Analysis},
  volume={13},
  year={2009},
  publisher={Elsevier}
}

@ARTICLE{8705271,
  author={Aubert, B. and Vazquez, C. and Cresson, T. and Parent, S. and de Guise, J. A.},
  journal={IEEE Transactions on Medical Imaging}, 
  title={Toward Automated 3D Spine Reconstruction from Biplanar Radiographs Using CNN for Statistical Spine Model Fitting}, 
  year={2019},
  volume={38},
  }

@article{hosseinian20153d,
  title={3D Reconstruction from Multi-View Medical X-ray images--review and evaluation of existing methods},
  author={Hosseinian, S and Arefi, H},
  journal={The International Archives of the Photogrammetry, Remote Sensing and Spatial Information Sciences},
  volume={XL-1/W5},
  year={2015},
  publisher={Copernicus GmbH}
}

@article{baka20112d,
  title={2D--3D shape reconstruction of the distal femur from stereo X-ray imaging using statistical shape models},
  author={Baka, Nora and Kaptein, Bart L and de Bruijne, Marleen and van Walsum, Theo and Giphart, JE and Niessen, Wiro J and Lelieveldt, Boudewijn PF},
  journal={Medical Image Analysis},
  volume={15},
  year={2011},
  publisher={Elsevier}
}

@article{karade20153d,
  title={3D femur model reconstruction from biplane X-ray images: a novel method based on Laplacian surface deformation},
  author={Karade, Vikas and Ravi, Bhallamudi},
  journal={International Journal of Computer Assisted Radiology and Surgery},
  volume={10},
  year={2015},
  publisher={Springer}
}

@article{andersen1984simultaneous,
  title={Simultaneous algebraic reconstruction technique (SART): a superior implementation of the ART algorithm},
  author={Andersen, Anders H and Kak, Avinash C},
  journal={Ultrasonic Imaging},
  volume={6},
  year={1984},
  publisher={Elsevier}
}

@inproceedings{arsigny2006log,
  title={A log-euclidean framework for statistics on diffeomorphisms},
  author={Arsigny, Vincent and Commowick, Olivier and Pennec, Xavier and Ayache, Nicholas},
  booktitle={International Conference on Medical Image Computing and Computer-Assisted Intervention (MICCAI)},
  year={2006},
}

@article{wasserthalTotalSegmentatorRobustSegmentation2023,
	title = {{TotalSegmentator}: {Robust} segmentation of 104 anatomic structures in {CT} images},
	volume = {5},
	journal = {Radiology: Artificial Intelligence},
	author = {Wasserthal, Jakob and Breit, Hanns-Christian and Meyer, Manfred T. and Pradella, Maurice and Hinck, Daniel and Sauter, Alexander W. and Heye, Tobias and Boll, Daniel T. and Cyriac, Joshy and Yang, Shan and Bach, Michael and Segeroth, Martin},
	year = {2023},
}

@article{tustisonANTsXEcosystemQuantitative2021,
	title = {The {ANTsX} ecosystem for quantitative biological and medical imaging},
	volume = {11},
	journal = {Scientific Reports},
	author = {Tustison, Nicholas J. and Cook, Philip A. and Holbrook, Andrew J. and Johnson, Hans J. and Muschelli, John and Devenyi, Gabriel A. and Duda, Jeffrey T. and Das, Sandhitsu R. and Cullen, Nicholas C. and Gillen, Daniel L. and Yassa, Michael A. and Stone, James R. and Gee, James C. and Avants, Brian B.},
	year = {2021},

}

@inproceedings{guedon2023sugar,
  title={Sugar: Surface-aligned gaussian splatting for efficient 3d mesh reconstruction and high-quality mesh rendering},
  author={Gu{\'e}don, Antoine and Lepetit, Vincent},
  booktitle={IEEE/CVF Conference on Computer Vision and Pattern Recognition (CVPR)},
  year={2024},
}

@InProceedings{10.1007/978-3-031-43999-5_66,
author="Cafaro, Alexandre
and Spinat, Quentin
and Leroy, Amaury
and Maury, Pauline
and Munoz, Alexandre
and Beldjoudi, Guillaume
and Robert, Charlotte
and Deutsch, Eric
and Gr{\'e}goire, Vincent
and Lepetit, Vincent
and Paragios, Nikos",
editor="Greenspan, Hayit
and Madabhushi, Anant
and Mousavi, Parvin
and Salcudean, Septimiu
and Duncan, James
and Syeda-Mahmood, Tanveer
and Taylor, Russell",
title="X2Vision: 3D CT Reconstruction from Biplanar X-Rays with Deep Structure Prior",
booktitle="International Conference on Medical Image Computing and Computer-Assisted Intervention (MICCAI)",
year="2023",
}

@article{van2022deep,
  title={Deep learning-based 2D/3D registration of an atlas to biplanar X-ray images},
  author={Van Houtte, Jeroen and Audenaert, Emmanuel and Zheng, Guoyan and Sijbers, Jan},
  journal={International Journal of Computer Assisted Radiology and Surgery},
  volume={17},
  year={2022},
  publisher={Springer}
}
